# Multimodal Sensor-Based Semantic 3D Mapping for a Large-Scale Environment

Jongmin Jeong[a], Tae Sung Yoon[b], and Jin Bae Park[a,*]

[a]*The School of Electrical and Electronic Engineering, Yonsei University, Seoul, 03722, Korea*
[b]*Department of Electrical Engineering, Changwon National University, Changwon-si, 51140, Korea*

## Abstract

Semantic 3D mapping is one of the most important fields in robotics, and has been used in many applications, such as robot navigation, surveillance, and virtual reality. In general, semantic 3D mapping is mainly composed of 3D reconstruction and semantic segmentation. As these technologies evolve, there has been great progress in semantic 3D mapping in recent years. Furthermore, the number of robotic applications requiring semantic information in 3D mapping to perform high-level tasks has increased, and many studies on semantic 3D mapping have been published. Existing methods use a camera for both 3D reconstruction and semantic segmentation. However, this is not suitable for large-scale environments and has the disadvantage of high computational complexity. To address this problem, we propose a multimodal sensor-based semantic 3D mapping system using a 3D Lidar combined with a camera. In this study, we build a 3D map by estimating odometry based on a global positioning system (GPS) and an inertial measurement unit (IMU), and use the latest 2D convolutional neural network (CNN) for semantic segmentation. To build a semantic 3D map, we integrate the 3D map with semantic information by using coordinate transformation and Bayes' update scheme. In order to improve the semantic 3D map, we propose a 3D refinement process to correct wrongly segmented voxels and remove traces of moving vehicles in the 3D map. Through experiments on challenging sequences, we demonstrate that our method outperforms state-of-the-art methods in terms of accuracy and intersection over union (IoU). Thus, our method can be used for various applications that require semantic information in 3D map.

## 1. Introduction

The inclusion of semantic information within a 3D map is becoming increasingly important in many fields, especially robotics (Rogers, 2013; Kostavelis, 2016). Existing 3D maps only contain geometry, which limits the robot's functionality to perform various tasks. To enable a robot to recognize an environment and act accordingly in a real 3D world, the robot needs to infer not only the geometry but also semantic information of surrounding environment. A semantic 3D map contains both geometry information and semantic information, allowing a robot to perform high-level tasks within the semantic 3D map.

In general, semantic 3D mapping involves geometric 3D mapping (for building a 3D map) and semantic segmentation (for obtaining semantic information); a semantic 3D map is produced by combining these. In recent years, studies on semantic 3D mapping have focused on camera-based systems. In (Sunderhauf, 2017), 3D reconstruction was performed by ORB-SLAM, and semantic information was extracted through a deep learning-based single-shot detector which used an RGB-D camera to build semantic 3D maps. In (Vineet, 2015), an incremental 3D reconstruction was carried out using a stereo camera and a CRF was used for semantic segmentation. Furthermore, in (Yang, 2017), a 3D reconstruction was produced from a stereo camera and high order conditional random fields (CRFs) were used for semantic labeling. McCormac et al. performed visual simultaneous localization and mapping (SLAM) with ElasticFusion and a CNN-based semantic segmentation method(McCormac, 2017). Several other studies used similar SLAM and CNN-based approaches (Cheng, 2017; Li, 2016). However, existing approaches have focused on a camera-based method for semantic 3D mapping, which have a limitation for applying to large-scale environments. Furthermore, visual SLAM cannot be performed in featureless environments.

In this study, we build a semantic 3D map with five labels: *road*, *sidewalk*, *vehicle*, *building*, and *vegetation*. The reason we consider five labels is that objects corresponding to these labels occupy most of the urban environments. The odometry of our system is estimated by integrating a GPS with an IMU, and the 3D map is generated by registering point clouds obtained from a 3D Lidar. For semantic segmentation, we use the CNN model to obtain the distributions for five labels. Following this, pixel label distributions on image are transferred to 3D grid space through coordinate transformation and Bayesian update schemes. Then, a 3D refinement process is performed to correct misclassified labels and remove traces of moving vehicles, thereby producing an accurate semantic 3D map. In the 3D refinement process, the segmentation accuracy is improved by using a 2D count grid representation and Bayes' rule, and moving vehicles are removed by combining the spatial context and the clustering method.

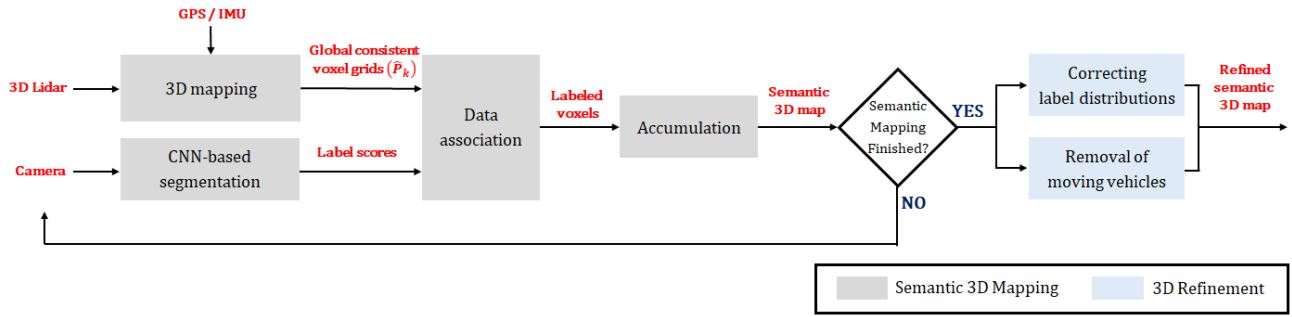

Figure 1: Flowchart of our semantic 3D mapping method.

In all, our main contributions are as follows:

- We proposed a multimodal sensor-based semantic 3D mapping system for large-scale environments.

- We demonstrated that 3D segmentation accuracy is improved against the state-of-the-art methods using the KITTI dataset.

- We developed a 3D refinement process to correct label distributions and remove traces of moving vehicles.

The remainder of this paper is organized as follows: Related work is reviewed in Section 2. Section 3 describes the proposed method in detail. Section 4 describes experiments that compare the proposed method against other state-of-the-art algorithms, and analyzes experimental results. Conclusions are presented in Section 5.

## 2. Related works

Semantic mapping is the process of attaching semantic information to a map. It uses 3D reconstruction as a tool to build a 3D map, and semantic segmentation to obtain semantic information. In this section, we briefly review other relevant studies. Geometric 3D mapping can be classified into three categories depending on the sensor for localization. The first is the visual SLAM, which uses a camera for localization and mapping. A remarkable early visual SLAM system was presented by Davison et al. (Davison, 2007). This system could estimate 3D trajectory of a monocular camera. Furthermore, in (Sthmer, 2010; Newcombe, 2011), researchers correctly estimated depth using a monocular camera, but were hampered by the scalability issues owing to memory requirements. In recent years, many open source algorithms, such as ElasticFusion (Whelan, 2015) and ORB-SLAM (Mur-Artal, 2017) have become widely available. Nonetheless, these camera-based approaches are not applicable in featureless environment, and it is difficult to build a 3D map in large-scale environments. The second 3D mapping category is the Lidar-based SLAM method. Bosse and Zlot et al. used a 2-axis Lidar to produce a point cloud which was registered by matching the geometric structures of local point clusters (Bosse, 2009). In (Zhang, 2014), researchers developed Lidar odometry and mapping (LOAM) method which estimates odometry and builds a 3D map by using a 3D Lidar; however, Lidar-based methods are difficult to apply in environments where there are no structural features. Finally, odometry is estimated by integrating the GPS with the IMU, and the 3D map is generated by accumulating point clouds measured from the Lidar. This is the most commonly used method for generating large-scale 3D maps for autonomous vehicles (Puente, 2013). It is the most suitable method for large-scale environments because of short computation time. In this paper, we choose this method for building a 3D map.

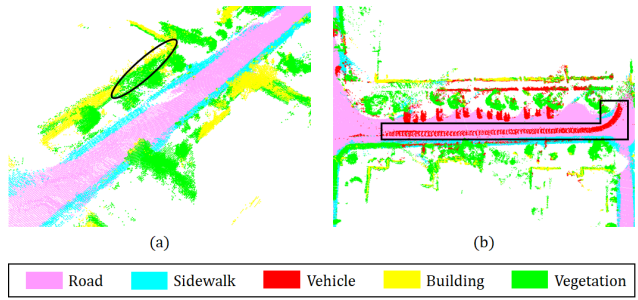

| Road | Sidewalk | Vehicle | Building | Vegetation |

Figure 2: Drawbacks of semantic 3D map: (a) example of wrong segmentation and (b) example of remaining traces

Many studies have been conducted on semantic segmentation, and many researchers solved this problem with Markov Random fields (MRFs) (Shotton, 2006) and CRFs (Krhenbhl, 2011). In recent years, studies have addressed semantic labeling problems by using CNNs. Long et al. introduced transposed convolutional layer and developed fully convolutional networks for segmentation (Long, 2015). In (Badrinarayanan, 2015), an encoder-decoder architecture with max unpooling layer and transposed convolutional layer was proposed. The cutting-edge method, namely, DilatedNet, achieved state-of-the-art performance on semantic segmentation by introducing atrous convolution, which extends the size of the receptive field without loss of resolution (Yu, 2015). Lin et al. developed RefineNet that receives information from different resolutions via long-range connections, thereby being able to obtain fine and coarse results (Lin, 2016). This method showed the best performance among state-of-the-art algorithms. Semantic 3D mapping requires fine and smooth semantic segmentation results; however, CNN-based methods still have limitations for producing these results.

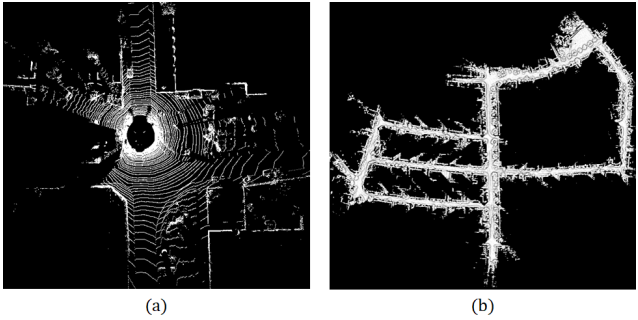

Figure 3: 3D mapping: (a) incoming point cloud and (b) generated dense 3D map.

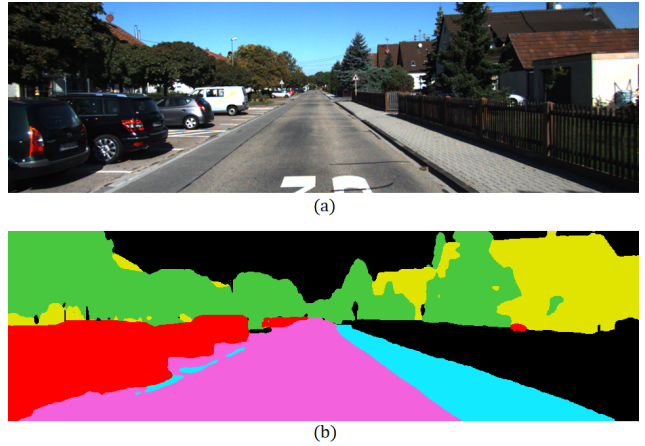

Figure 4: 2D semantic segmentation: (a) incoming image and (b) prediction.

Although many works on semantic 3D mapping exist, they have focused on camera-based methods. In (Cheng, 2017), ORB-SLAM was applied for building a 3D map while semantic segmentation was performed through CRF-RNN, and both were integrated to generate a semantic 3D map. Hermans et al. proposed a novel label transfer method based on Bayesian updates and dense pairwise 3D CRFs to transfer labels from 2D to 3D (Hermans, 2014). In (Sengupta, 2013), a straightforward solution was proposed to directly transfer 2D image labels to 3D by using back-projection; however, it took considerable time to run outdoors. Furthermore, Sengupta et al. proposed a volumetric labeling approach, which simultaneously inferred the 3D structure and object labeling (Sengupta, 2015). Vineet et al. produced an incremental 3D reconstruction from stereo pairs and used a random forest with a CRF for the semantic segmentation (Vineet, 2015). In (Li, 2016), the semantic 3D mapping problem was solved by combining deep learning and semi-dense SLAM based on a monocular camera. Finally, in (Yang, 2017), semantic 3D maps were built using a CRF model with higher order. Nonetheless, these works are not suitable for large-scale or featureless environments since they utilized camera for SLAM.

## 3. Approach

### 3.1. System overview

Our semantic 3D mapping method uses a 3D Lidar and a camera as its main sensors and focuses on large-scale environments (particularly, urban environments). As illustrated in Fig. 1, our method mainly consists of building and refining a semantic 3D map. To build the semantic 3D map, geometric reconstruction is performed using the 3D Lidar to generate a globally consistent point cloud for each frame. In parallel, the CNN for the semantic segmentation takes an image as input and returns a set of per-pixel label probabilities. Following this, data association is performed based on the Bayesian update rule. This enables us to keep track of the label distributions based on the CNN's prediction and determines the final label of each voxel in the 3D map. The semantic 3D map is built frame-by-frame; however, there are some limitations. First, buildings and vegetation are misclassified each other owing to inaccurate performance of the CNN and projection errors, as shown in Fig. 2(a). Second, traces of moving vehicles remain on the 3D map as shown in Fig. 2(b), owing to the presence of vehicles that move concurrently with our platform. To solve these problems, the 3D refinement process is applied to built semantic 3D map. In the 3D refinement process, estimation of labels based on the 2D count grid representation is performed to solve misclassification. Furthermore, any traces of moving vehicles are removed by combining the spatial context and the clustering method.

### 3.2. Semantic 3D mapping

#### 3.2.1. 3D mapping

3D mapping is a fundamental part of our framework (as previously mentioned, our map is built using a 3D Lidar), To build a large-scale 3D map in real-time, we estimate the odometry for each frame by combining a GPS with an IMU. This is an ideal method for large-scale environments owing to its simple computation process and short computation time. In our 3D mapping system, the odometry is composed of a three-dimensional translation vector and a three-dimensional rotation matrix, and it is estimated every frame. Each point cloud is transformed using the estimated odometry for global 3D map generation as shown in (1):

$$\hat{P}_i = T_i P_i \qquad (1)$$

where $T_i = \begin{bmatrix} R_i & t_i \\ 0 & 1 \end{bmatrix}$, $R_i$ is the estimated rotation matrix, $t_i$ is the estimated translation vector, and $P_i$ is the point cloud of $i$th frame. For memory and computational efficiency, transformed point clouds are changed to voxel grids. Each grid stores occupancy status, and the grid size is set to $0.2m \times 0.2m \times 0.2m$, which is sufficient for generating dense 3D map in a large-scale environment. As transformed voxel grids are accumulated for entire frames as (2), dense 3D map is generated as shown in Fig. 3.

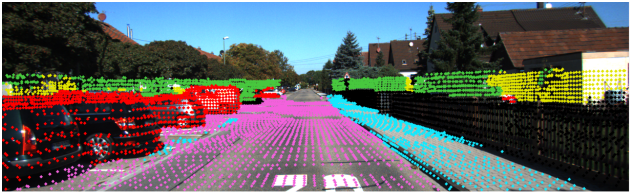

Figure 5: Alignment of coordinate systems for projecting semantic information onto voxel grids. Colored points depict labeled voxels.

$$M_{1:k} = \{\hat{P}_i^v, \quad i = 1, 2, ...k\} \tag{2}$$

where $\hat{P}_i^v$ is the voxel set obtained from $i$th point cloud, and $M_{1:k}$ represents the 3D map from $I$st frame to $k$th frame.

### 3.2.2. Semantic segmentation

The goal of 2D semantic segmentation is to assign the correct label for each pixel in an image. Along with the success of CNNs in classification (Krizhevsky, 2012), CNN-based approaches have shown remarkable improvements in various computer vision applications. In 2D semantic segmentation, many successful CNN architectures have been developed (Long, 2015; Badrinarayanan, 2015; Yu, 2015; Paszke, 2016; Lin, 2016). In this work, we chose RefineNet as our semantic segmentation method. RefineNet exploits multi-level feature maps by using residual connections to generate a high-resolution semantic feature map. The reason we choose the RefineNet is that it is the open source and has the best performance compared to state-of-the-art methods (Lin, 2016). We used the network trained on the Cityscape dataset for our work, and fine-tuned the network using the Camvid dataset. To generate a semantic 3D map on urban environments, we selected five labels(*road*, *sidewalk*, *vehicle*, *building*, and *vegetation*). Hence, the generated semantic 3D map also contains information on these labels. In our framework, the semantic segmentation returns scores over five labels at each pixel for data association. These scores had a size of $W \times H \times 5$, where $W$ and $H$ are the width and height of the input image, respectively. Furthermore, $5$ represents the number of labels. Fig. 4 shows the results of semantic segmentation.

### 3.2.3. Data association

The data association step determines the label of each voxel. To achieve this, both globally consistent voxels (generated by 3D mapping) and label distributions (obtained by 2D semantic segmentation) are combined; this combination is performed using two-stage pipelines. First, to align between a 3D Lidar coordinate system and a camera coordinate system, the 3D Lidar coordinate system is transformed to the camera coordinate system using extrinsic parameter which represents the positional relationship between two sensors. Following this, a label is assigned to each voxel by using the recursive Bayesian update rule which estimates the label distributions of a 3D map. Transformation from the 3D Lidar coordinate system to the camera

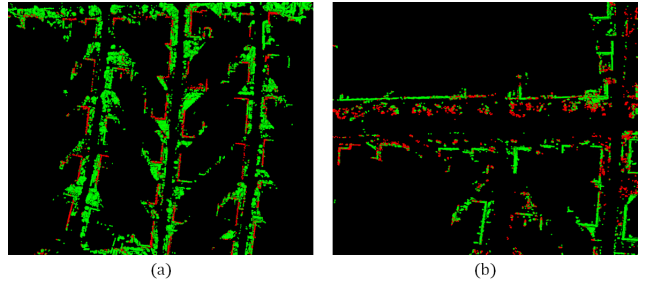

Figure 6: Results of correcting label distributions: (a)*vegetation* and (b)*building*. Points removed by the process are shown in red for better visualization.

coordinate system is formulated as shown in (3) and (4).

$$\begin{bmatrix} u \\ v \\ w \end{bmatrix} = P_{L2C} \begin{bmatrix} x_v \\ y_v \\ z_v \\ 1 \end{bmatrix} \tag{3}$$

$$\begin{bmatrix} u_c \\ v_c \end{bmatrix} = \begin{bmatrix} \dfrac{u}{w} \\ \dfrac{v}{w} \end{bmatrix} \tag{4}$$

where $(x_v, y_v, z_v)$ are the $x, y, z$ value of the voxel, respectively, $P_{L2C}$ is the transformation matrix which converts the 3D Lidar coordinate system into the camera coordinate system. $u_c$ and $v_c$ are the camera pixel coordinates corresponding to $(x_v, y_v, z_v)$. After aligning the two coordinate systems, the semantic information is projected onto the corresponding voxels, as shown in Fig. 5, thereby building the semantic 3D map. This is formulated as shown in (5).

$$L_v(x_v, y_v, z_v) = I_S(u_c, v_c) \tag{5}$$

where $I_S$ is a result image of semantic segmentation, and $L_v(\cdot)$ means the label of the voxel, $(x_v, y_v, z_v)$. However, since each voxel could be observed as a different label in each frame, it is necessary to fuse label observations. For the label fusion, the Bayesian update rule is applied, as shown in (6).

$$p(l_k^v|I_{1:k}, P_{1:k}) \approx \frac{1}{Z} p(l_k^v|I_k, P_k) p(l_{k-1}^v|I_{1:k-1}, P_{1:k-1}) \tag{6}$$

where $Z$ is the normalization constant and $l_k^v$ denotes the labels of voxel $v$ at time $k$. $I_{1:k-1}$ and $P_{1:k-1}$ are the images and point clouds to $k-1$, respectively. In equation (6), the label probability distribution is simply multiplication followed by normalization to update for each incoming image and point cloud. As a result, the final label of each voxel is estimated by maximizing label probability distribution as shown in (7)

$$L_v(v) = \operatorname*{argmax}_{l^v} p(l^v|I, P) \tag{7}$$

where $v$ is a voxel, and $l^v$ means labels which can be *road*, *sidewalk*, *vehicle*, *building* and *vegetation*. After label fusion, we can produce the semantic 3D map where each voxel contains only one label information.

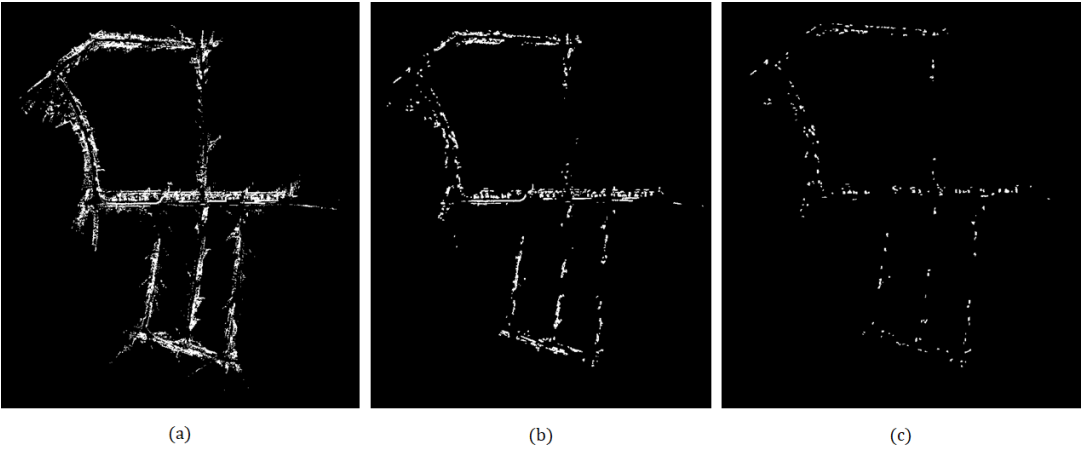

Figure 7: Results of moving vehicle removal: (a)*vehicle* 3D map, (b)after applying spatial context and (c)after clustering method

### 3.3. 3D refinement

There are several drawbacks to the built semantic 3D map. First, results obtained from 2D semantic segmentation include misclassified pixels, and the precise outline of objects cannot be found. Also, inaccurate extrinsic parameter leads to projection errors. Consequently, some voxels with misclassification are included in the semantic 3D map, especially in *building* and *vegetation* categories, as shown in Fig. 2(a). Second, traces of moving vehicles remain in the semantic 3D map. To correct these, the 3D refinement process is performed for the entire semantic 3D map in a batch. This process consists of two steps. The first step corrects the label distributions by using a 2D count grid representation in which each grid stores the number of voxels belonging to the grid. The second step is to remove traces and noises by using the spatial context and the clustering method.

#### 3.3.1. Correcting label distributions

In our semantic 3D map, some voxels corresponding to *building* and *vegetation* are wrongly segmented owing to projection errors, while voxels classified as *road*, *sidewalk*, and *vehicle* are segmented correctly. In this section, our goal is to correct labels of misclassified voxels in *building* and *vegetation*. Through analysis of voxels in *building* and *vegetation*, it is verified that most of voxels in *building* and *vegetation* are correctly segmented, and sparse voxels are misclassified as shown in Fig. 2(a). To take advantage of this feature, the 2D count grid representation is newly defined to correct errors of label distributions. The 2D count grid representation is formulated as (8).

$$G_c^l(x, y) = \sum_z M^l(x, y, z) \tag{8}$$

where $M^l$ denotes the 3D map corresponding to the label $l$ that can be extracted from the semantic 3D map, and $M^l$ has a value of 1 if the grid is occupied. $G_c^l$ means 2D count grids for the label $l$. To modify labels of 3D voxels in *building* and *vegetation*, labels of 2D grids are inferred using count grids, which is based on Bayes' rule. It is formulated as (9).

$$p(l^{2D}|G_c^B, G_c^V) = \frac{1}{Z} p(l^{2D}) p(G_c^B, G_c^V | l^{2D}) \tag{9}$$

where $l^{2D}$ is the label of 2D grids which can be either *building* or *vegetation*. $G_c^B$ and $G_c^V$ are 2D count grids of *building* and *vegetation*, respectively. The label of each 2D grid can be determined by maximizing the label probability distribution as shown in (10).

$$L_{2D} = \underset{l^{2D}}{\arg\max} \, p(l^{2D}|G_c^B, G_c^V) \tag{10}$$

Furthermore, in order to reflect 2D grids' labels into 3D voxels, we use the spatial assumption that voxels with the same $x$ and $y$ values are likely to have the same label. That is, labels of voxels in *building* and *vegetation* are corrected as shown in (11).

$$L_v(x_v, y_v, z_v) = L_{2D}(x_v, y_v) \tag{11}$$

Using this process, we modify the label distributions for *building* and *vegetation*. As shown in Fig. 6, the voxels corresponding to *building* are removed from *vegetation* set, and the voxels corresponding to *vegetation* are removed from *building* set.

#### 3.3.2. Moving vehicle removal

As shown in the built semantic 3D map, there are two categories in *vehicle* label. The first is parked vehicles, which are static. These vehicles do not require removal. The second is moving vehicles. They remain as traces or noises in the 3D map and are needed to be removed for clarity. In addition, since semantic segmentation cannot accurately detect the contour of vehicles, few voxels are wrongly projected to adjacent objects such as buildings or vegetation. To remove these voxels, the spatial assumption that vehicles must be on the road is applied. Because road is well-segmented in semantic 3D map, we could erase voxels that are not on the road, and it is expressed by (12).

$$M^V(x, y, z) = \begin{cases} 1, & (x, y) \in M^R \\ 0, & \text{otherwise} \end{cases} \tag{12}$$

where $M^V$ and $M^R$ are the 3D maps corresponding to *vehicle* and *road*, respectively. After this process, we can obtain a set of voxels corresponding to static vehicles or moving vehicles. Then, a clustering method is applied to group nearby voxels together. Among the numerous methods of clustering, we chose density-based spatial clustering of applications with noise (DB-SCAN) algorithm due to its low parametric characteristics (Ester, 1996). The groups corresponding to moving vehicles are of greater length and have more data than the groups of static vehicles. Thus, we could remove the groups of moving vehicles by using the number of data and length as constraints as shown in (13).

$$C_i = \begin{cases} static, & \text{if } (D_i < \eta_D) \wedge (L_i < \eta_L) \\ moving, & \text{otherwise} \end{cases} \tag{13}$$

where $C_i$ is a status of $i$th group, $D_i$ and $L_i$ are the number of data and length of $i$th group, respectively. $\eta_D$ and $\eta_L$ are the threshold values. This process leaves only static vehicles in the semantic 3D map as shown in Fig. 7.

## 4. Experimental results

We conducted experiments on three sequences of the KITTI dataset (Geiger, 2013), which is publicly available. The sequences were recorded in urban environments including *road*, *sidewalk*, *vehicle*, *building* and *vegetation*. Our system was evaluated on these sequences and compared with state-of-the-art methods such as (Yang, 2017), (Vineet, 2015),(Sengupta, 2013), and (Sengupta, 2015). The experiments were implemented with MATLAB, on an Intel Core i7 with 3.40GHz and NVIDIA GeForce GTX 1080Ti.

### 4.1. Dataset

We evaluated our method using the KITTI dataset, which contains a variety of outdoor sequences. All sequences in the KITTI dataset were recorded using a vehicle which was equipped with a 3D Lidar, stereo cameras, a GPS, and an IMU. The KITTI dataset is challenging as it contains various objects, moving vehicles, and changes in lighting conditions. We chose the 15th, 18th, and 27th sequences for the demonstration because these sequences effectively describe urban environments. The 15th sequence is a sequence of road environment with a duration of 30 seconds and a total of 303 frames. The 18th sequence is a long sequence set in an urban environment, with a duration of 276 seconds and a total of 2769 frames. Finally, the 27th sequence is a set in a road environment, with a duration of 111 seconds and a total of 1112 frames.

### 4.2. Implementation details

To build the 3D map, the odometry was estimated by integrating the data of the GPS and IMU which were mounted on the roof of a car. Furthermore, the data of a 3D Lidar (Velodyne HDL-64E) was used for 3D mapping process. A voxel grid representation was applied and the grid size was set to $0.2m \times 0.2m \times 0.2m$ for memory efficiency. For the semantic segmentation, the data of a front color camera mounted on the roof of the car was used. The segmentation was performed for five labels (*road*, *sidewalk*, *vehicle*, *building*, and *vegetation*), and we chose RefineNet as our semantic segmentation tool. In the data association process, calibration parameters provided in the KITTI dataset were used for coordinate transformation. To remove traces of moving vehicles, $\eta_D$ and $\eta_L$ were set to 1500 and 6$m$, respectively. These values were determined empirically through numerous experiments.

### 4.3. Qualitative evaluation

To objectively evaluate our multimodal sensor-based semantic 3D mapping algorithm, three challenging sequences were used. The qualitative results of our algorithm are presented in Fig. 8, which shows one top view of the whole sequence and three close-up views for each sequence. Our system was able to reconstruct the surrounding environment and assign the correct label for each voxel in challenging sequences well. Moreover, experimental results showed that objects with low height (such as *road* and *sidewalk*) were accurately segmented without applying a 3D refinement process, as projection errors were less likely to occur. On the other hand, projection errors were occurred with objects of greater height (such as *building*, *vehicle*, and *vegetation*), which occasionally lead to incorrect segmentation results, as shown in Fig. 2(a). Furthermore, vehicles that move leave traces, which make the map messy as shown in Fig. 2(b). To solve these problems, the 3D refinement process was performed in a batch. Fig. 9 shows the results with and without the 3D refinement process, respectively, and it demonstrates the advantages of this process. As shown in Fig. 9, before the 3D refinement process, there are not only some voxels which are wrongly segmented but also some voxels corresponding to traces of moving vehicles. It is difficult to solve them only from 2D images. However, it is possible to reassign labels correctly and remove traces of moving vehicles effectively by using the 3D refinement process. In addition, our multimodal sensor-based 3D semantic mapping achieved good performance in large-scale environments.

### 4.4. Quantitative evaluation

In this section, we quantitatively evaluate and compare the 3D segmentation accuracy with state-of-the-art methods. To objectively evaluate our algorithm, we selected (Yang, 2017), (Vineet, 2015), (Sengupta, 2013) and (Sengupta, 2015) as comparisons because these algorithms have shown reliable performance in semantic 3D mapping for urban environments. However, it is important to note that these are all camera-based methods, and are different from our multimodal sensor-based method. We adopted the standard metric of label accuracy and intersection over union (IoU) to evaluate the performance of our 3D segmentation method. These metrics are defined as follows:

$$\text{Accuracy} = \frac{\text{TP}}{\text{TP} + \text{FP}} \tag{14}$$

$$\text{IoU} = \frac{\text{TP}}{\text{TP} + \text{FP} + \text{FN}} \tag{15}$$

Table 1: Quantitative results for our 3D segmentation approach on the KITTI dataset. The bold fonts indicate the best results.

| | Method | Road | Sidewalk | Vehicle | Building | Vegetation | Average |
|---|---|---|---|---|---|---|---|
| Accuracy | Yang (Yang, 2017) | 98.7 | 93.8 | 95.5 | 98.2 | **98.7** | 96.9 |
| | Vineet(Vineet, 2015) | 98.7 | 91.8 | 94.1 | 97.2 | 94.1 | 95.1 |
| | Sengupta(Sengupta, 2013) | 97.8 | 86.5 | 88.5 | 96.1 | 86.9 | 91.1 |
| | Sengupta(Sengupta, 2015) | 97.0 | 73.4 | 72.5 | 89.1 | 81.2 | 82.6 |
| | Ours | **99.3** | **98.8** | **98.4** | **98.9** | 97.3 | **98.5** |
| IoU | Yang (Yang, 2017) | 96.6 | 90.0 | 94.6 | 95.4 | 91.0 | 93.5 |
| | Vineet(Vineet, 2015) | 94.7 | 73.8 | 79.5 | 88.3 | 83.2 | 83.9 |
| | Sengupta(Sengupta, 2013) | 96.3 | 68.4 | 63.5 | 83.8 | 74.3 | 77.2 |
| | Sengupta(Sengupta, 2015) | 87.8 | 49.1 | 55.8 | 73.8 | 65.2 | 66.3 |
| | Ours | **98.4** | **96.9** | **96.8** | **97.4** | **96.3** | **97.1** |

where TP, FP and FN represent True Positive, False Positive and False Negative, respectively. Comparison results are presented in Table 1; these results were obtained by conducting experiments on the KITTI dataset. As mentioned in Section 1, we segmented on five classes that are majority of urban environments. From the Table 1, our method greatly outperforms other semantic mapping algorithms in all categories. For the class *vehicle*, there was a significant accuracy increase of 3% over the state-of-the-art methods, and for the *building*, we achieved 0.7% improvement. Overall, the average accuracy was increased by 1.6%. In terms of IoU, there was a performance improvement of 2.2% for *vehicle*, and 2% for *building*. For *vegetation*, there was an increase of 5.3%. There are two main reasons why our method outperforms existing methods. First, we used the latest 2D CNN as a semantic segmentation tool. RefineNet which we used has proved to be better performance than other semantic segmentation methods. Second, we developed a new 3D refinement process which improves accuracy for the *vehicle*, *building* and *vegetation* classes. Furthermore, we would like to highlight the scalability of our semantic 3D mapping pipelines. Because our algorithm has a small amount of computation compared with the camera-based approaches, it is suitable for large-scale environments.

## 5. Conclusion

In this paper, a multimodal sensor-based semantic 3D mapping algorithm that uses a 3D Lidar for 3D mapping and a camera for semantic segmentation is proposed. Existing works have focused only on camera-based algorithms for semantic 3D mapping, which are difficult to apply in large-scale environments as well as environments with few features. Furthermore, these methods require a large amount of computation. To overcome these difficulties, a simple method which estimates the odometry by integrating a GPS and an IMU is used to build a 3D map. This method is suitable for both large-scale and featureless environments. In addition, the latest 2D CNN method is adopted for semantic segmentation. Following this, the semantic information is combined with the 3D map by using coordinate system transformation and a Bayesian update scheme.

Finally, a novel 3D refinement process is developed to reduce errors caused by misprojection and moving vehicles. Comparisons with state-of-the-art algorithms on challenging sequences demonstrate that our algorithm outperforms others in terms of label accuracy and IoU.

Our paper offers many compelling avenues for future work. One area of interest that we would like to explore is to build the indoor semantic 3D maps. To achieve it, a Lidar-based SLAM should be studied. A semantic SLAM method which SLAM and semantics benefit each other could also an interesting area worthy of future study.

## Acknowledgment

This research was supported by a National Research Foundation of Korea (NRF) grant funded by the Korea Government (MSIP) (No. NRF-2015R1A2A2A01007545), and was partially supported by Microsoft Research.

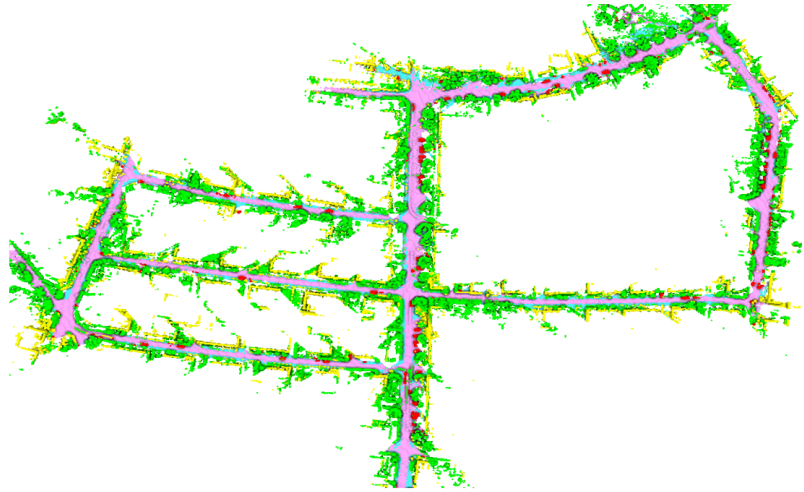

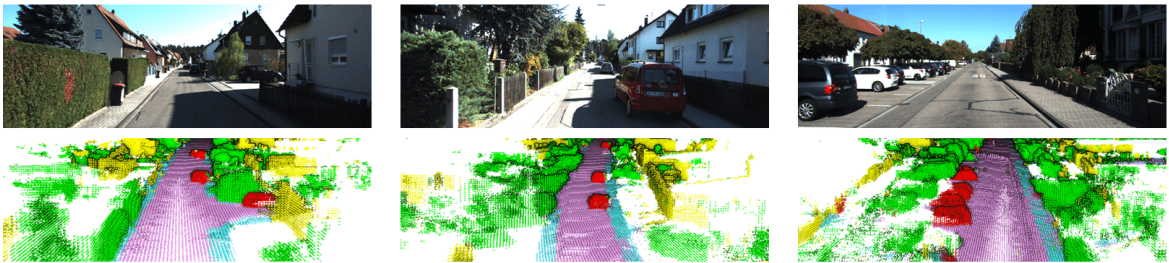

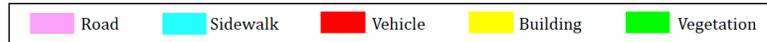

| Road | Sidewalk | Vehicle | Building | Vegetation |

(a) KITTI sequence 18

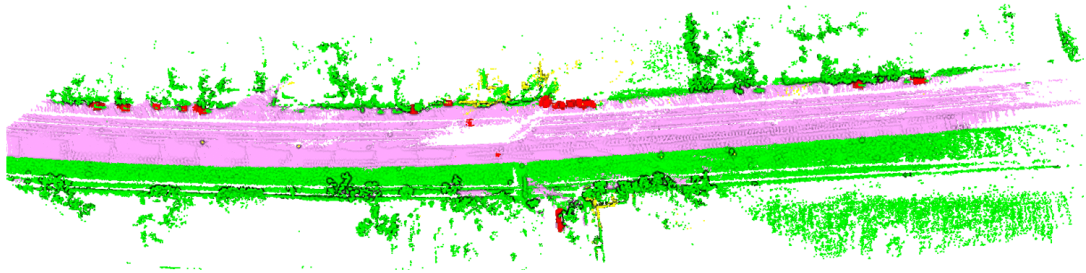

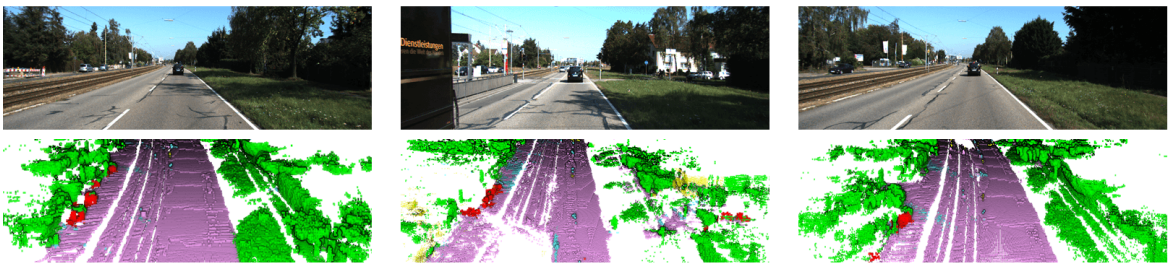

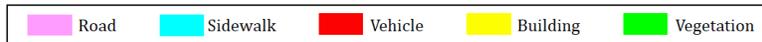

| Road | Sidewalk | Vehicle | Building | Vegetation |

(b) KITTI sequence 15

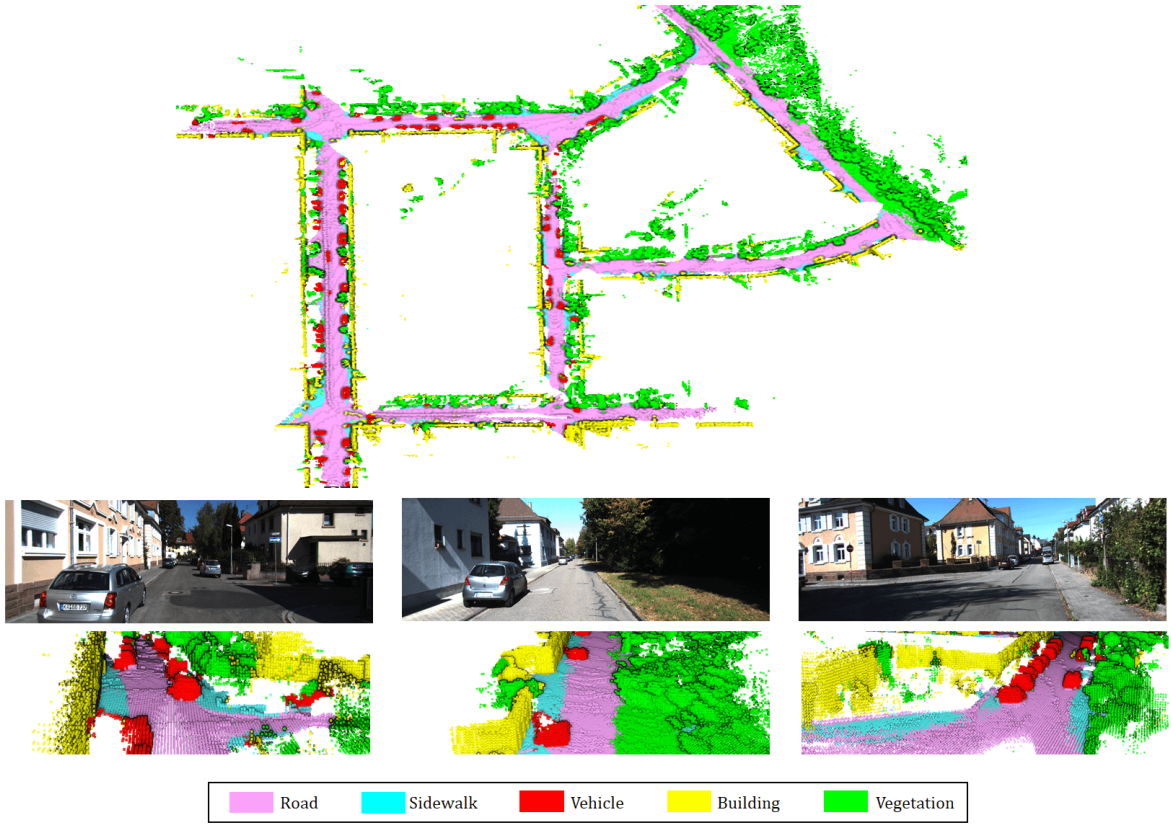

| Road | Sidewalk | Vehicle | Building | Vegetation |

(c) KITTI sequence 27

Figure 8: Visualization of semantic 3D mapping. Top view of the whole sequence and close-up views of semantic 3D map. It demonstrates that our method works well in large-scale environments.

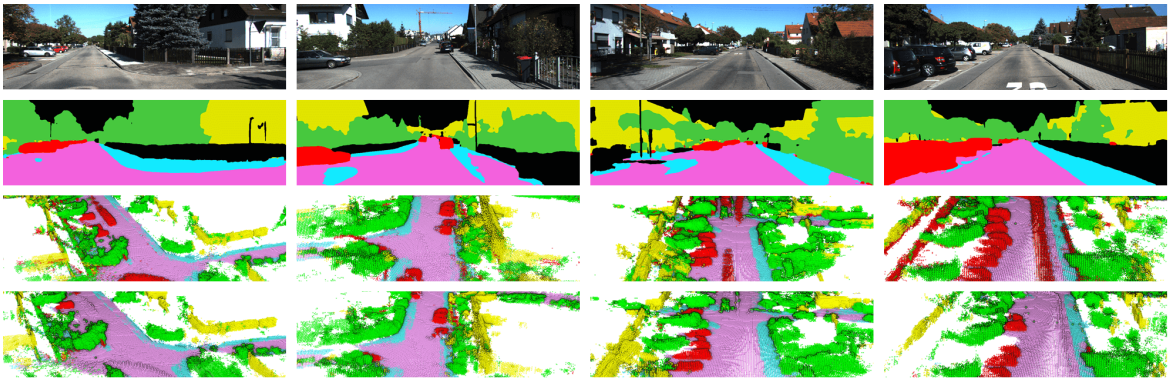

Figure 9: Qualitative results. (First Row) original images. (Second Row) 2D semantic segmentation. (Third Row) semantic 3D map without 3D refinement. (Bottom Row) semantic 3D map with 3D refinement